\documentclass{hcj}

\usepackage{graphicx}
\usepackage{color}
\usepackage[table]{xcolor}
\usepackage{natbib}

\usepackage{array} 

\usepackage{rotating} 

\usepackage[misc]{ifsym} 

\definecolor{myred}{rgb}{.8,.0,.0}

\definecolor{mygreen}{rgb}{.0,.8,.0}

\newcommand{\wontfix}[1]{}

\newcommand{\numPapersTotal}{57}

\newcommand{\accessedDate}{Feb.~2019}
\newcommand{\urlfootnote}[1]{\footnote{\url{#1} Accessed \accessedDate}}

\newcommand{\pctTaskClassification}{42}
\newcommand{\pctTaskSegmentation}{39}
\newcommand{\pctTaskSandC}{12}
\newcommand{\pctTaskOther}{7}

\newcommand{\pctApplicationBrain}{9}
\newcommand{\pctApplicationEye}{14}
\newcommand{\pctApplicationLung}{9}
\newcommand{\pctApplicationBreast}{0}
\newcommand{\pctApplicationHeart}{2}
\newcommand{\pctApplicationAbdomen}{23}
\newcommand{\pctApplicationHisto}{28}
\newcommand{\pctApplicationMultiple}{7}
\newcommand{\pctApplicationOther}{9}

\newcommand{\pctInteractionRateImage}{52}
\newcommand{\pctInteractionRateNew}{90}
\newcommand{\pctInteractionRateExisting}{14}
\newcommand{\pctInteractionDrawShape}{38}
\newcommand{\pctInteractionDrawNew}{90}
\newcommand{\pctInteractionDrawExisting}{14}
\newcommand{\pctInteractionClick}{25}
\newcommand{\pctInteractionCompare}{9}
\newcommand{\numInteractionCompare}{5}

\newcommand{\pctPlatformCommercial}{55}
\newcommand{\pctPlatformCustom}{23}
\newcommand{\pctPlatformVolunteer}{10}
\newcommand{\pctPlatformOther}{8}
\newcommand{\pctPlatformProtoSim}{5}
\newcommand{\pctRewardLow}{34}
\newcommand{\pctRewardUnknown}{34}
\newcommand{\pctRewardZero}{24}
\newcommand{\pctRewardHigh}{5}
\newcommand{\pctRewardHour}{3}
\newcommand{\pctAnnotatorsSingle}{5}
\newcommand{\pctAnnotatorsMultiple}{63}
\newcommand{\pctAnnotatorsUnknown}{33}
\newcommand{\pctSmallAndMediumScale}{72}
\newcommand{\pctVerySmallScale}{5}
\newcommand{\pctLargeScale}{20}
\newcommand{\pctUnknownScale}{3}
\newcommand{\pctMediumRangeAnnotationsFromMultiple}{66}

\newcommand{\pctPreprocess}{86} 
\newcommand{\pctFilterPast}{16} 
\newcommand{\pctFilterTestBefore}{12} 
\newcommand{\pctFilterTestDuring}{23} 
\newcommand{\pctFilterGold}{23} 
\newcommand{\pctAggregateMajority}{23} 
\newcommand{\pctAggregateWeighted}{16} 
\newcommand{\pctAggregateNummerical}{12} 
\newcommand{\pctAggregate}{61} 
\newcommand{\pctComparedToGold}{79} 
\newcommand{\pctIndirectComparison}{16} 
\newcommand{\pctNoComparison}{5} 
\newcommand{\pctGoldStandardSingle}{25} 
\newcommand{\pctGoldStandardMultiple}{37} 
\newcommand{\pctGoldStandardSameTask}{40} 
\newcommand{\pctGoldStandardDifferentTask}{40} 
\newcommand{\pctGoldStandardUnknownTask}{19} 
\newcommand{\propGoldStandardMultipleVariation}{3 of 21} 

\journalyear{}
\journalvolume{}
\journalissue{}
\journalpages{}
\journalcopyright{}
\articledoi{}

\begin{document}
\title{A Survey of Crowdsourcing in\\ Medical Image Analysis}

\author{Silas N. {{\O}}rting$^{\textrm{\Letter}}$\affil{University of Copenhagen, Copenhagen, Denmark}
  \and Andrew Doyle$^*$\affil{McGill Centre for Integrative Neuroscience, Montreal, Canada}
  \and Arno van Hilten$^*$\affil{Erasmus Medical Center, Rotterdam, The Netherlands}
  \and Matthias Hirth$^*$\affil{Technische Universit\"at Ilmenau, Ilmenau, Germany}
  \and Oana Inel$^*$\affil{Vrije Universiteit Amsterdam, Amsterdam, The Netherlands}\affil{Delft University of Technology, Delft, The Netherlands}
  \and Christopher R. Madan$^*$\affil{University of Nottingham, Nottingham, United Kingdom}
  \and Panagiotis Mavridis$^*$\affil{Delft University of Technology, Delft, The Netherlands}
  \and Helen Spiers$^*$\affil{University of Oxford, Oxford, United Kingdom}\affil{Zooniverse, University of Oxford, Oxford}
  \and Veronika Cheplygina$^{\textrm{\Letter}}$\affil{Eindhoven University of Technology, Eindhoven, The Netherlands}
  }

\authorrunning{S. N. {{\O}}rting, A. Doyle, A. van Hilten, M. Hirth, O. Inel, C. R. Madan, P. Mavridis, H. Spiers and V. Cheplygina}

\maketitle
\emph{$^{\textrm{\Letter}}$ silas@di.ku.dk, v.cheplygina@tue.nl}\\
\emph{* These authors contributed equally and are listed alphabetically by last name}

\clearpage

\begin{abstract}
Rapid advances in image processing capabilities have been seen across many domains, fostered by the  application of machine learning algorithms to ``big-data''. However, within the realm of medical image analysis, advances have been curtailed, in part, due to the limited availability of large-scale, well-annotated datasets. One of the main reasons for this is the high cost often associated with producing large amounts of high-quality meta-data. Recently, there has been growing interest in the application of crowdsourcing for this purpose; a technique that has proven effective for creating large-scale datasets across a range of disciplines, from computer vision to astrophysics. Despite the growing popularity of this approach, there has not yet been a comprehensive literature review to provide guidance to researchers considering using crowdsourcing methodologies in their own medical imaging analysis. In this survey, we review studies applying crowdsourcing to the analysis of medical images, published prior to July 2018 \wontfix{Do we want to extend this with more recent papers?}. We identify common approaches, challenges and considerations, providing guidance of utility to researchers adopting this approach. Finally, we discuss future opportunities for development within this emerging domain. 
 \end{abstract}

\section{Introduction} \label{sec:introduction}
The limited availability and size of labeled datasets for training machine learning algorithms is a common problem in medical image analysis~\citep{greenspan2016guest,litjens2017survey,cheplygina2018supervised}. In several other fields, crowdsourcing - defined as the outsourcing of tasks to a crowd of individuals~\citep{howe2006rise}- has been found effective for labeling large quantities of data. For example, in computer vision crowdsourcing has been used to annotate large datasets of images and videos with various tags~\citep{kovashka2016crowdsourcing}. 

Due to the success of crowdsourcing, several researchers have recently applied these techniques to the annotation of medical images. Although such images present specific challenges, including absence of expertise of the crowd, several early papers such as \citep{mitry2013crowdsourcing,mavandadi2012distributed,maier2014can} have demonstrated promising results. Despite the growing interest, there has not been an overview of the work in this field. In this paper we summarize existing literature on crowdsourcing in medical imaging.

\begin{figure}
    \centering
    \includegraphics[width=\columnwidth]{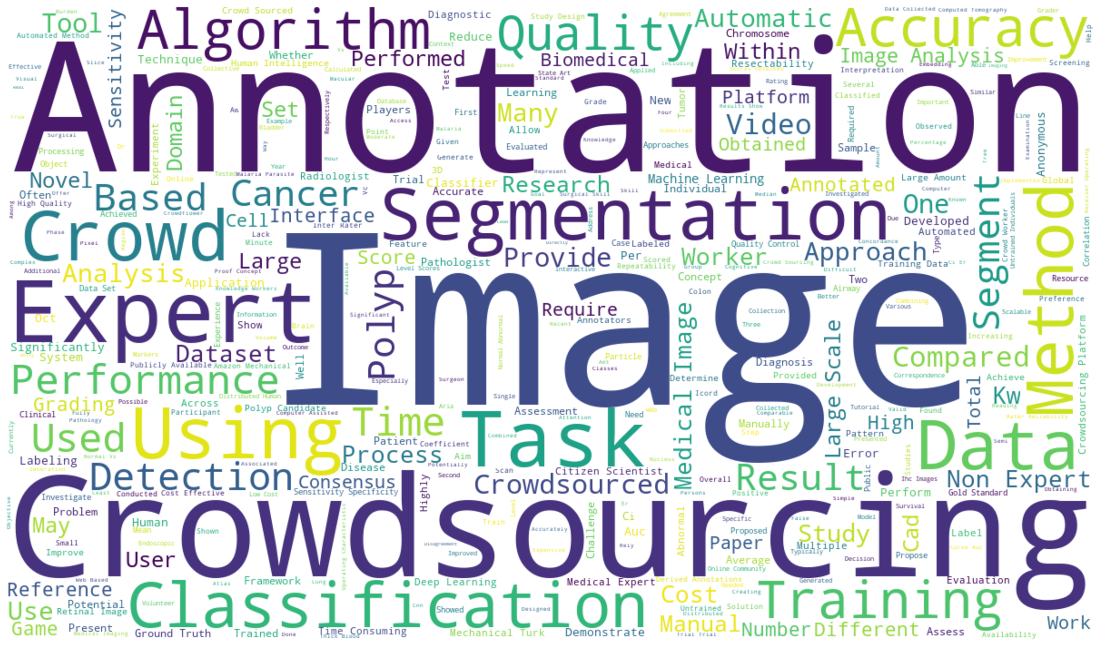}
    \caption{A word cloud of the abstracts of the surveyed papers.}
    \label{fig:wordcloud}
\end{figure}

This paper originated during the Lorentz workshop ``Crowdsourcing in medical image analysis'' in June 2018\urlfootnote{https://www.lorentzcenter.nl/lc/web/2018/967/info.php3?wsid=967&venue=Snellius}. As participants of the workshop, we searched Google Scholar with the query ``crowdsourcing AND (medical or biomedical)'' and screened the results for papers focusing on the topic. Google Scholar was selected due to previous papers highlighting the poor indexing of the topic in databases and the high prevalence of crowdsourcing papers in conferences \citep{wazny2017crowdsourcing}. Additional papers were identified for inclusion by examining the references and citations of selected papers. We only included papers where the crowd was involved in the analysis of medical or biomedical images, for example by annotating them. Our search strategy resulted in \numPapersTotal~papers. Key terms emerging from these studies are illustrated in Fig.~\ref{fig:wordcloud}. Five key dimensions were identified for discussion;  the application involved, the type of interaction between the crowd and the images, the scale of the task (such as the number of images), the type of evaluation performed on the crowd annotations, and the results of the evaluation. 

There are a number of surveys which are related to this work. However, they are quite different in scope: 
\begin{itemize}
    \item \cite{ranard2014crowdsourcing} survey crowdsourcing in health and medical research. They identify four tasks: problem solving, data processing, monitoring and surveying and cover 21 papers published until March 2013. In contrast, we only focus on papers where image analysis (i.e. data processing) is involved. 

    \item \cite{kovashka2016crowdsourcing} survey crowdsourcing in computer vision. The surveyed papers focus on analysis of everyday/natural images. Only one of the 195 referenced papers (\cite{gurari2015how}) uses biomedical data. 
    
    \item \cite{wazny2017crowdsourcing} present a meta-review of crowdsourcing from 2006 to 2016. Similar to \cite{ranard2014crowdsourcing}, they take a more broad view of crowdsourcing. They review 48 existing review papers until August 2015, focusing on how each review categorizes the papers, for example by platform, size of crowd, and so forth.

    \item \cite{alialy2018review} is most similar to our survey, but only focuses on crowdsourcing in human pathology. They do a systematic search with several steps, excluding conference papers or abstracts, and summarize seven papers. The coverage of literature is therefore much more limited than in this work. 
\end{itemize}

The paper is organized as follows. The five dimensions according to which we analyzed the papers are described in more detail in Sections~\ref{sec:applications} to~\ref{sec:results}. We then discuss overall trends, limitations, and opportunities for future research in Section~\ref{sec:discussion}.

\section{Applications} \label{sec:applications}
There are a variety of crowdsourcing applications to medical imaging data addressed in the papers surveyed in this work. We group these applications by (1) the type of task performed by the crowd, (2) the biomedical content of the image and (3) the dimensionality of the images.

\subsection{Type of task}

An important task in medical image analysis is classification, and \pctTaskClassification\% of the surveyed papers focus on this task. Classification can refer to assigning a label to an entire image, such as diagnosing whether a chest CT image contains any abnormalities. Classification can also refer to assigning a label to a part of the image, for example, the type of abnormality located in a particular region of interest. Other types of labels include non-diagnostic labels such as image modality~\citep{herrera2014crowdsourcing}, visual attributes~\citep{cheplygina2018crowd}, and assessing the quality of the image~\citep{keshavan2018combining}. These three types of labels are based more on visual characteristics, and thus might be easier to provide than diagnostic labels without any medical training.  

A further \pctTaskSegmentation\% of the papers focus on localization or segmentation. Typically the goal is to delineate the boundary of an entire healthy structure, or of an abnormality such as a lesion. The difference with how we define the classification task above is that instead of providing information about the image, the annotator has to modify the image, by providing positions or outlines. These tasks rely more on visual characteristics than classification tasks, and may be more easily explained to a non-expert crowd. 

In \pctTaskSandC\% of the papers both classification and segmentation are addressed. Often this means that the annotator first has to indicate if the structure of interest is visible, and if yes, to locate it in the image. 

Finally, \pctTaskOther\% of papers request less standard tasks from their crowd. For example, \cite{maier2015crowdtruth} focuses on determining correspondence between pairs of images. Although this is a type of detection task, where the annotator has to locate points of interest in an image, it is also different since a point of reference is already provided. Another example is \cite{orting2017crowdsourced}, where the annotator has to decide which image is more similar to a reference image. This is a type of classification problem, but again relying more on visual features than on prior knowledge.

\subsection{Type of image}

Medical images are acquired at vastly different scales and locations depending on the physiological measurement of interest. The imaging acquisition modality and strategy depends heavily on the scale of the anatomy of interest, and different technologies' expected contrast with surrounding tissues. Here we categorize the images by where in the body the image originates from, which narrows down the modality. We use the following categorization, also used in two recent surveys of medical imaging \citep{litjens2017survey,cheplygina2018supervised}: brain, eye, heart, breast, lung, abdomen, histology/microscopy, multiple, other. 

We compare the distribution of applications surveyed in this work with the two other surveys in Table~\ref{tab:comparison_surveys}. An interesting observation is that \cite{litjens2017survey} and \cite{cheplygina2018supervised} have a similar distribution of applications despite surveying different topics: \citeauthor{litjens2017survey} covers deep learning, where a larger dataset is preferred, while \citeauthor{cheplygina2018supervised} covers weakly supervised learning, where datasets are smaller in size. Given that crowdsourcing is often proposed as an alternative to weakly supervised learning, it is surprising that the current survey has a different distribution of papers.

\begin{table}
    \centering
\begin{tabular}%
  {>{\raggedright\arraybackslash}p{1.2cm}%
   >{\raggedleft\arraybackslash}p{1.7cm}%
   >{\raggedleft\arraybackslash}p{2cm}%
   >{\raggedleft\arraybackslash}p{1.7cm}%
  }
        Application & This survey & \cite{cheplygina2018supervised} & \cite{litjens2017survey} \\ \hline
         Brain & \pctApplicationBrain\% & 21\%  & 18\%\\
         Eye & \pctApplicationEye\% & 4\% & 5\% \\
         Lung & \pctApplicationLung\% & 13\% & 14\% \\
         Breast & \pctApplicationBreast\%  & 6\% & 7\% \\
         Heart & \pctApplicationHeart\%  & 4\% & 7\% \\
         Abdomen & \pctApplicationAbdomen\% & 14\% & 9\% \\
         Histo/Micro & \pctApplicationHisto\%  & 17\% & 20\% \\
         Multiple & \pctApplicationMultiple\% & 12\% & 4\% \\
         Other & \pctApplicationOther\% & 10\% & 16\% \\ \hline                           
    \end{tabular}
    \caption{Comparison of the distribution of applications in this survey and two other recent surveys in medical image analysis. Percentages are rounded to the nearest whole number.}
    \label{tab:comparison_surveys}
\end{table}

Many of the papers in this survey are aimed at 2D images. The most common application is histopathology/microscopy with \pctApplicationHisto\% of all the papers, followed by retinal images with \pctApplicationEye\% of the papers. Both applications are over-represented compared to \citep{litjens2017survey} and \cite{cheplygina2018supervised}. This overrepresentation in crowdsourcing studies may be because many retinal and microscopic images are acquired in 2D, which might be easier to use in a crowdsourcing study than 3D images.

Breast and heart images, which were already not well represented in the other two surveys, are almost absent in crowdsourcing studies. Both applications can be aimed at 2D or 3D images. However, perhaps due to lack of datasets or perceived difficulty of assessing these images, these applications are almost never considered for crowdsourcing. 

Several other papers address applications where images are often 3D, such as the brain (\pctApplicationBrain\%) and the lungs (\pctApplicationLung\%). Compared to \citep{litjens2017survey,cheplygina2018supervised}, brain and lung images are underrepresented in crowdsourcing. This could be due to complexity of images or limitations in interfaces. One approach for dealing with 3D images is to select 2D parts of the original 3D images. For example, \citep{orting2017crowdsourced,oneil2017crowdsourcing} select axial slices. \citep{cheplygina2016early} shows patches of 2D projections in various directions in the image. Others circumvent the 3D problem by presenting a video to the users where the entire image is displayed as a sequence of 2D frames~\citep{boorboor2018crowdsourcing}. Only a few of the papers addressing 3D images, present images in 3D~\citep{huang2017swiftree,sonabend2017defining}.

The last type of data that is addressed is video, common for endoscopy and colonoscopy (both in the abdomen category). Several different approaches are used for presenting video data: 2D frames~\citep{maier2014crowdsourcing,maier2015crowdtruth,maier2016crowd,heim2018large,bittel2017how}, 3D renderings~\citep{nguyen2012distributed,mckenna2012strategies}, short video clips~\citep{park2017crowdsourcing}, or longer videos that can be paused and annotated~\citep{park2018crowd}.

Other applications of crowdsourcing include segmenting hip joints in 2D MRI~\citep{chavez2013crowdsourcing}, rating visual characteristics of dermatological images~\citep{cheplygina2018crowd} and assessing surgical performance~\citep{malpani2015study,holst2015crowd}. Two papers \citep{foncubierta2012ground,herrera2014crowdsourcing} look at multiple applications, where the task is classifying image modality, rather than segmentation or diagnosis. A few papers address segmentation in multiple modalities: \citep{gurari2016investigating} focus on both natural and biomedical images, \citep{lejeune2017expected} address segmentation across four medical applications.

\begin{table*}
  \rowcolors{2}{gray!25}{white}
\centering
\begin{sideways}
\resizebox{!}{7.5cm}{
  \begin{tabular}{l | l l | l | l l l | l l l l} \\
    & \multicolumn{2}{c}{Section~\ref{sec:applications}}    & \multicolumn{1}{c}{Section~\ref{sec:interaction}}    & \multicolumn{3}{c}{Section~\ref{sec:scale}}    & \multicolumn{4}{c}{Section~\ref{sec:evaluation}} \\
    Paper                              & Task     & Domain   & Interaction     & I  & Platform  & Reward     & Filtering     & Aggregation       & Comparison & Gold standard   \\
    \hline
    \cite{albarqouni2016aggnet}        & classify & histo    & rate            & M  & custom    & unknown    & before/during & majority/weighted & indirect   & multiple experts \\
    \cite{albarqouni2016playsourcing}  & other    & histo    & rate            & L  & custom    & volunteers & none          & weighted          & direct     & multiple experts \\
    \cite{bittel2017how}               & segment  & abdomen  & draw            & L  & none      & none       & none          & none              & na         & na               \\
    \cite{boorboor2018crowdsourcing}   & segment  & lung     & draw            & S  & paid      & low        & during        & none              & direct     & multiple experts \\
    \cite{brady2014rapid}              & classify & eye      & rate            & S  & paid      & unknown    & none          & none              & direct     & ?                \\
    \cite{brady2017improving}          & classify & eye      & rate            & M  & paid      & low        & none          & majority/weighted & direct     & ?                \\
    \cite{bruggemann2018exploring}     & segment  & histo    & click           & L  & paid      & low        & none          & none              & direct     & multiple experts \\
    \cite{cabrera2017counting}         & segment  & histo    & click           & XS & volunteer & volunteers & none          & other             & direct     & ?                \\
    \cite{chavez2013crowdsourcing}     & segment  & other    & draw            & M  & custom    & volunteers & after         & none              & direct     & other            \\
    \cite{cheplygina2016early}         & segment  & lung     & draw            & S  & paid      & unknown    & after         & average           & direct     & one expert       \\
    \cite{cheplygina2018crowd}         & classify & other    & rate            & S  & students  & volunteers & none          & average           & indirect   & ?                \\
    \cite{della2014preliminary}        & s+c      & histo    & click           & S  & paid      & unknown    & during        & average           & direct     & one expert       \\
    \cite{dos2015crowdsourcing}        & classify & histo    & rate            & L  & volunteer & volunteers & during        & average           & direct     & one expert       \\
    \cite{eickhoff2014crowd}           & classify & histo    & rate            & M  & paid      & low        & none          & majority          & direct     & one expert       \\
    \cite{foncubierta2012ground}       & classify & multiple & rate            & L  & paid      & unknown    & during        & none              & direct     & one expert       \\
    \cite{ganz2017crowdsourcing}       & segment  & brain    & draw            & S  & paid      & low        & none          & average           & direct     & one expert       \\
    \cite{gur2017towards}              & classify & heart    & rate            & M  & custom    & unknown    & none          & none              & indirect   & multiple experts \\
    \cite{gurari2015howtocollect}	   & segment  &	other	 & draw	           & M	& paid	    & low	     & before/during & majority/weighted & direct     &	multiple experts \\ 
    \cite{gurari2016investigating}     & s+c      & multiple & rate+draw       & M  & paid      & low        & before/after  & majority          & direct     & multiple experts \\
    \cite{heim2018large}               & s+c      & abdomen  & rate+draw       & M  & paid      & low        & before        & majority/weighted & direct     & multiple experts \\
    \cite{heller2017web}               & segment  & abdomen  & draw            & XS & custom    & unknown    & none          & none              & direct     & other            \\
    \cite{herrera2014crowdsourcing}    & classify & multiple & rate            & L  & paid      & volunteers & none          & none              & indirect   & other            \\
    \cite{holst2015crowd}              & classify & other    & rate            & S  & paid      & low        & before        & other             & direct     & multiple experts \\
    \cite{huang2017swiftree}           & classify & lung     & click           & ?  & custom    & unknown    & none          & other             & direct     & ?                \\
    \cite{irshad2015crowdsourcing}     & segment  & histo    & click+draw      & M  & paid      & unknown    & before/during & none              & direct     & multiple experts \\
    \cite{irshad2017crowdsourcing}     & s+c      & histo    & rate+draw       & M  & paid      & unknown    & before/during & majority/weighted & direct     & one expert       \\
    \cite{keshavan2018combining}       & classify & brain    & rate            & M  & custom    & volunteers & during        & weighted          & direct     & multiple experts \\
    \cite{lawson2017crowdsourcing}     & classify & histo    & click           & S  & volunteer & hourly     & none          & none              & direct     & multiple experts \\
    \cite{lee2014mechanical}           & segment  & eye      & draw            & S  & paid      & low        & none          & none              & direct     & na               \\
    \cite{lee2016use}                  & segment  & eye      & draw            & S  & paid      & low        & before        & none              & direct     & na               \\
    \cite{leifman2015leveraging}       & s+c      & eye      & rate+draw+click & ?  & custom    & volunteers & during        & other             & direct     & multiple experts \\
    \cite{lejeune2017expected}         & segment  & multiple & click           & S  & experts   & unknown    & none          & average           & indirect   & ?                \\
    \cite{luengo2012crowdsourcing}     & segment  & histo    & click           & S  & volunteer & volunteers & during        & other             & direct     & multiple experts \\
    \cite{maier2014can}                & segment  & abdomen  & draw            & S  & paid      & unknown    & none          & none              & indirect   & one expert       \\
    \cite{maier2014crowdsourcing}      & other    & abdomen  & click+compare   & S  & paid      & unknown    & none          & other             & direct     & multiple experts \\
    \cite{maier2015crowdtruth}         & other    & abdomen  & click+compare   & S  & paid      & low        & none          & other             & direct     & ?                \\
    \cite{maier2016crowd}              & segment  & abdomen  & click+compare   & M  & paid      & unknown    & none          & majority          & indirect   & ?                \\
    \cite{malpani2015study}            & classify & other    & rate+compare    & M  & ?         & hourly     & during        & majority/weighted & direct     & multiple experts \\
    \cite{mavandadi2012distributed}    & classify & histo    & rate            & M  & none      & unknown    & before/during & none              & direct     & one expert       \\
    \cite{mckenna2012strategies}       & classify & abdomen  & rate            & M  & paid      & low        & before        & other             & direct     & ?                \\
    \cite{mitry2013crowdsourcing}      & classify & eye      & rate            & S  & paid      & low        & none          & none              & direct     & multiple experts \\
    \cite{mitry2015crowdsourcing}      & classify & eye      & rate            & M  & paid      & low        & before/after  & none              & direct     & multiple experts \\
    \cite{mitry2016accuracy}           & s+c      & eye      & rate+draw       & S  & paid      & low        & before/after  & majority          & direct     & multiple experts \\
    \cite{nguyen2012distributed}       & classify & abdomen  & rate            & M  & paid      & low        & before        & majority          & direct     & ?                \\
    \cite{oneil2017crowdsourcing}      & segment  & lung     & draw            & S  & custom    & volunteers & after         & majority          & direct     & one expert       \\
    \cite{orting2017crowdsourced}      & other    & lung     & compare         & M  & paid      & low        & before        & other             & indirect   & multiple experts \\
    \cite{park2016c2a}                 & classify & abdomen  & rate            & M  & paid      & low        & after         & majority/weighted & direct     & one expert       \\ 
    \cite{park2017crowdsourcing}       & classify & abdomen  & rate            & M  & paid      & unknown    & none          & majority          & direct     & one expert       \\
    \cite{park2018crowd}               & segment  & abdomen  & click           & M  & paid      & unknown    & before/during & none              & direct     & one expert       \\
    \cite{rajchl2016learning}          & segment  & brain    & click           & L  & custom    & volunteers & none          & none              & indirect   & one expert       \\
    \cite{rajchl2017employing}         & segment  & abdomen  & ?               & M  & none      & none       & none          & none              & na         & na               \\
    \cite{sameki2016icord}             & segment  & histo    & draw            & M  & paid      & low        & none          & other             & direct     & ?                \\
    \cite{sharma2017crowdsourcing}     & segment  & histo    & draw            & S  & paid      & low        & none          & other             & direct     & na               \\
    \cite{smittenaar2018harnessing}    & classify & histo    & rate            & L  & custom    & volunteers & none          & weighted          & direct     & multiple experts \\
    \cite{sonabend2017defining}        & classify & brain    & rate            & ?  & experts   & unknown    & none          & none              & direct     & na               \\
    \cite{sullivan2018deep}            & classify & histo    & rate+draw       & L  & custom    & volunteers & none          & other             & direct     & ?                \\
    \cite{timmermans2016crowdsourcing} & s+c      & brain    & rate+draw       & M  & custom    & volunteers & none          & none              & na         & na               \\
  \end{tabular}
  }
  \end{sideways}
  \caption{Overview of papers with the task (s+c = segment and classify), application, interaction, number of images (XS=$<10$, S=10 to 100, M=100 to 1000, L=$>1000$), platform, reward, filtering, aggregation, type of comparison used (na = not applicable), source of gold standard. A question mark ``?'' indicates the information was not found in the paper.}
  \label{tab:literature}
\end{table*}

\section{Interaction} \label{sec:interaction}
An important aspect of crowdsourcing medical image annotations is task design. The interplay between the type of image data, the type of annotations that are needed and the available tools for annotation, needs to be considered to successfully crowdsource annotations. A major component of task design is choosing how workers interact with the task. The type of interaction influences time per task and the required level of expertise and training, which ultimately translates into cost and quality. 
We identified four categories of interaction tasks across the studies surveyed:
\begin{itemize}
  \item Rate an entire image
  \item Draw shapes to identify regions of interest
  \item Click on specific locations
  \item Compare two or more images
\end{itemize}

Furthermore, we also observed that studies generally had crowds either (1) create entirely new annotations on unlabeled data, or (2) make responses based on pre-existing annotations, e.g., output from automated segmentation methods.

Rating entire images was the most common interaction and was the main task of \pctInteractionRateImage\% of the studies surveyed here. Ratings took many forms, identifying the presence/absence of specific visual features \citep{sonabend2017defining}, counting number of cells \citep{smittenaar2018harnessing}, assessing intensity of cell staining \citep{dos2015crowdsourcing}, or discriminating healthy samples from diseased \citep{mavandadi2012distributed}.
Most commonly, crowd workers were asked to create new annotations (\pctInteractionRateNew\% of rating tasks). Less commonly, crowd workers were asked to validate pre-existing annotations (\pctInteractionRateExisting\%). One study involved both validating pre-existing annotations and creating new ones \citep{heim2018large}, so the percentages do not sum to 100\%. Existing annotations were the output of automated methods \citep{bittel2017how, ganz2017crowdsourcing, gur2017towards} half of the time, and the crowdsourced annotations were used to identify instances with errors to be corrected.

Drawing a shape was the second most common task, comprising \pctInteractionDrawShape\% of studies. Here crowd workers were asked to draw bounding boxes or segment outlines of structures of interest. Sometimes, this was only after identifying if a structure was present in the image or not \citep{heim2018large}. Similar to rating images, drawing shapes was used as an interaction for both creating new annotations (\pctInteractionDrawNew\% of drawing tasks) and validating existing annotations (\pctInteractionDrawExisting\%).
In the case of evaluating existing annotations, drawing was used as a means to indicate the location of errors in segmentations produced by automated methods \citep{bittel2017how, ganz2017crowdsourcing}.

Clicking on specific locations was the third most used interaction, occurring in \pctInteractionClick\% of studies. Clicking was only used to create new annotations such as identifying the precise location of specific cells, abnormalities, or artifacts within an image. The use of multiple clicks to outline a structure was considered a ``drawing a shape'' interaction. Selecting points was also used in pairs of video frames to determine the stereotactic correspondence of two video streams for follow-up 3D reconstruction \citep{maier2014crowdsourcing, maier2015crowdtruth, maier2016crowd}.

Comparing two or more images was the least used interaction, occurring in only \numInteractionCompare~(\pctInteractionCompare\%) of studies. In all cases, comparisons were used to create new annotations, such as marking corresponding points in two consecutive video frames \citep{maier2015crowdtruth,maier2016crowd} or to choose which of two images was more similar to a target image \citep{orting2017crowdsourced}.

Overall, crowds were more often used to create new annotations, than to make judgments on existing annotations, which was done only in \citep{bittel2017how,foncubierta2012ground,ganz2017crowdsourcing,gur2017towards,herrera2014crowdsourcing}. Ratings and drawing of shapes can be used to obtain more detailed annotations than information already present in datasets. Clicking interactions are sometimes used to identify specific image features, but more commonly used to create bounding boxes or draw object boundaries. Evaluating existing annotations is always done with rating or drawing interactions. \wontfix{maybe add citations here}

\section{Platform, Scale and Wages} \label{sec:scale}
In this section we summarize the main meta parameters and settings of crowdsourcing experiments.
First, we classify the reviewed papers based on the type of platform used to perform the crowdsourcing experiments.
Second, we report on the scale of the experiments where we consider 1) the number of images annotated and 2) the number of annotators per image.
Finally, we summarize the wages paid to crowd workers.

\subsection{Crowdsourcing platforms}
\label{subsec:crowdsourcing_platform}
A potentially important factor that varies across the surveyed papers is the choice of platform for conducting crowdsourcing experiments. We classify the platforms into six categories: paid commercial marketplaces such as Amazon Mechanical Turk\urlfootnote{https://www.mturk.com} and FigureEight\urlfootnote{https://www.figure-eight.com} (formerly known as CrowdFlower), volunteers such as Zooniverse\urlfootnote{https://www.zooniverse.org} and Volunteer Science\urlfootnote{https://volunteerscience.com}, custom recruitment/platforms, lab participants, experts and simulation or no experiment at all. The most common choice is a commercial platform (\pctPlatformCommercial\%). The second most common choice is a custom platform (\pctPlatformCustom\%) followed by a volunteer platform (\pctPlatformVolunteer\%). The remaining \pctPlatformOther\% were almost equally divided into the other categories with around \pctPlatformProtoSim\% of all papers reporting prototypes or simulation studies.

\subsection{Scale}
\label{subsec:crowdsourcing_experiment_scale}
We summarize the scale of the crowdsourcing experiments in terms of number of images annotated and number of annotations per image.

\subsubsection{Number of images}
We classify the number of images into four categories: very small (less than 10 annotated images), small (10 to 100 annotated images), medium (100 to 1000 annotated images) and large (more than 1000 annotated images). The large majority of reviewed papers, \pctSmallAndMediumScale\%, report small and medium scale experiments, while a smaller part report large experiments (\pctLargeScale\%) or very small experiments (\pctVerySmallScale\%). However, in around \pctUnknownScale\% of the reviewed papers, the scale of the experiments is not reported. 

\subsubsection{Number of annotations per image}
\label{subsec:annotations_per_unit}
 We divide the number of annotations per image into two categories: a single annotator per image (\pctAnnotatorsSingle\%) or multiple annotators per image (\pctAnnotatorsMultiple\%). Surprisingly, for \pctAnnotatorsUnknown\% of surveyed papers the number of annotations per image is not reported nor can it be inferred.

Overall, the experiments using a single annotator per image involve either simulations or locally recruited, volunteer-based annotators that are not remunerated. The number of annotators per image for experiments using multiple annotators per image ranges from 2 to 5000. However, the majority (\pctMediumRangeAnnotationsFromMultiple\%) of these experiments use between 5 to 25 annotators per image.

\subsection{Annotators Wage}
\label{subsec:annotators_reward}
We classify the wage given to annotators into six different categories: a few dollars per hour, less than or equal to \$0.10 per annotation, more than \$0.10 per annotation, volunteers (no monetary incentive), not specified (if we have no information about compensation) and none (if no actual experiment or recruitment took place). 

More than a third (\pctRewardUnknown\%) of papers did not specify anything about wage. In \pctRewardLow\% of papers the wage  was less than or equal to \$0.10, in \pctRewardZero\% of papers crowds where volunteers with no monetary incentive, in \pctRewardHigh\% of papers the wage  was more than \$0.10, and in \pctRewardHour\% of papers the wage  was an hourly payment of a few dollars per hour.

Overall, very few and mainly the papers that mention an hourly payment considered crowd worker wages in relation to minimum wage rules and regulations.

\section{Evaluation} \label{sec:evaluation}
In this section we describe how the crowdsourced annotations are evaluated. This is done via two strategies:
\begin{itemize}
    \item ensuring sufficient quality of annotations by preprocessing 
    \item estimating the utility of the crowd annotations for the task at hand
\end{itemize}

Although the two strategies are closely related and should be considered jointly when designing crowdsourcing experiments, it is informative to consider them separately here.

The first strategy is closely related to the field of quality control in crowdsourcing. Numerous approaches exist to tackle this, starting from simple majority voting and worker filtering to sophisticated statistical and machine learning methods that consider workers' specific skills, task difficulty and clarity of task descriptions.
The second strategy is more domain-specific, as different tasks may have different levels of tolerance for errors.

\subsection{Preprocessing of annotations}

Preprocessing of annotations broadly covers what is done to the crowdsourced annotations prior to using them for their intended purpose. It includes filtering individual annotations and/or aggregating annotations. Of the surveyed papers, \pctPreprocess\% perform some form of preprocessing.

\subsubsection{Filtering individuals}
One way to filter annotations, is to remove annotations made by ``poorly performing'' annotators. Most crowdsourcing platforms offer a rating score for workers that provides an estimate of their performance, based on their percentage of previously approved tasks. This score is used in \pctFilterPast\% of surveyed papers to filter workers prior to assigning tasks. A related approach, used in \pctFilterTestBefore\% of surveyed papers, is to exclude workers that fail a test task prior to the actual tasks. A refinement of this, used in \pctFilterTestDuring\% of surveyed papers, is to integrate separate test tasks in the tasks and exclude workers that fail the tests. One example is adding a smiley face to colonoscopy videos to ensure attention \citep{park2018crowd}.

Another common filtering approach for individual workers, used in \pctFilterGold\% of surveyed papers, is comparing annotations to gold standard annotations. In this case, tasks with known gold standard annotations, are injected into the regular working process. A worker's correspondence with the gold standard can then be used to estimate individual worker performance. In contrast to platform scores and unrelated test tasks, this approach assesses worker performance on the specific task, allowing more fine-grained worker selection.

\subsubsection{Aggregating results}
One of the main benefits of crowdsourcing is the fast and cost-effective collection of a large number of annotations. This allows aggregating annotations to reduce noise in the individual annotations. 

Majority voting is widely used due to its computational and conceptual simplicity and was found in \pctAggregateMajority\% of the papers. In the context of medical image analysis, majority voting is applied to annotations, ratings, and also to aggregate slices of images. One example is presented in \cite{heim2018large} where the authors used crowdsourcing for organ segmentation in computed tomography scan. Multiple organ outlines are collected via an online tool and pixel-wise majority voting is applied to improve the accuracy of the segmentation.

In the case of numerical ratings, mean and median statistics are also used in \pctAggregateNummerical\% of the papers to determine a final annotation. For example, ~\cite{cheplygina2016early} used the median to aggregate the areas of the annotations created by individual workers.

A more sophisticated version of the majority vote uses additional information about the general quality of workers. This information can be derived if workers perform multiple tasks or if gold standard data is available. Weighted voting is used in \pctAggregateWeighted\% of surveyed papers, for example in \citep{keshavan2018combining}, where the XGBoost algorithm was used to estimate annotator weights and in \citep{brady2017improving} where annotator weights are estimated as the probability an annotator is correct while taking task difficulty into account.

\subsection{Evaluating annotations}
Evaluating how well crowd annotations solve the intended purpose is most commonly (\pctComparedToGold\% of surveyed papers) done by directly comparing crowdsourced annotations to a gold standard. In about \pctIndirectComparison\% of surveyed papers crowd annotations are used for training a machine learning method, and the performance of the machine learning method used to indirectly evaluate annotations. The remaining \pctNoComparison\% have no evaluation of how well annotations solve the intended purpose.

The gold standard originates from different sources. 
In about \pctGoldStandardSingle\% of surveyed papers the gold standard is based on a single expert, in about \pctGoldStandardMultiple\% the gold standard is based on multiple experts and in the remaining papers the number of experts is not reported or no expert gold standard is used. Using a gold standard based on a single expert can be problematic since experts often disagree on all but the most trivial tasks. However, only \propGoldStandardMultipleVariation~papers that use multiple experts consider how well experts agree. 

Expert-based gold standards are generally not obtained from experts performing exactly the same task as the crowd. In several cases the only difference in expert and crowd tasks is due to differences in user interface, e.g. a clinical workstation for experts and a web interface for crowds. As long as the fundamental task is the same (e.g. count cells) and the user interface has not been dramatically changed we consider the expert and crowd tasks to be the same. Using this definition, about \pctGoldStandardSameTask\% of the papers use the same task and about \pctGoldStandardDifferentTask\% use a different task. In the remaining \pctGoldStandardUnknownTask\% it is either not reported or no expert gold standard is used.

There are several reasons for asking crowds to perform a different task than what experts have done for the gold standard.
Some papers use a simplified version of the expert task in order to make the task easier or more suitable as a small self-contained task. For example, ranking relative performance in pairs of surgical videos instead of grading performance in each \citep{malpani2015study}; assessing visual similarity of images instead of classifying disease patterns \citep{orting2017crowdsourced}; refining segmentation proposals instead of performing a full segmentation \citep{maier2016crowd}; annotating polyps in a single frame instead of in a full video \citep{park2017crowdsourcing} or counting stained cells instead of classifying disease status \citep{irshad2017crowdsourcing}.
Other papers focus on changing the user interface, such as in \citep{lejeune2017expected} where an eye tracker is used for segmentation instead of a mouse, or in \citep{albarqouni2016playsourcing,mavandadi2012distributed} where the user interface is changed to support gamification strategies.

In a few papers, evaluation is focused on variation in annotations. For example, in \citep{lee2014mechanical,lee2016use} where annotations are evaluated in terms of inter-rater reliability; and in \citep{heller2017web,huang2017swiftree,leifman2015leveraging,sonabend2017defining} where individual annotations are compared to aggregated annotations. Measuring variability of annotations it not directly useful for evaluating the correctness of annotations. However, annotation variability is essential when evaluating how much the crowdsourced annotations can be trusted. Additionally, variation provides an indirect measure of correctness. Large variation can indicate that annotations are often wrong, while small variation indicates that annotations are often correct or the task has been designed such that annotators are consistently wrong. \wontfix{References for these statements?}

\section{Results and recommendations} \label{sec:results}
Here, we provide an overview of the primary results and recommendations emerging from the papers examined in this review. Complementary to the topics discussed in Section~\ref{sec:evaluation} we consider (1) How effective is the application of crowdsourcing to medical image analysis? (2) Recommendations to ensure data quality. 

\subsection{How effective is the application of crowdsourcing to medical image analysis?}
The vast majority of studies examined in this review found crowdsourcing to be a valid approach for data production. Crowdsourcing of medical image analysis was noted to be an accurate approach \citep{lawson2017crowdsourcing}, that can produce large quantities of annotations needed to solve high-throughput problems requiring human input \citep{irshad2015crowdsourcing,dos2015crowdsourcing,lee2014mechanical,maier2014crowdsourcing}. Crowdsourcing can be used to create new annotations or make existing data more robust, both cheaper and faster than annotation by medical experts \citep{rajchl2016learning,holst2015crowd,gurari2016investigating,eickhoff2014crowd,park2017crowdsourcing}.

Although the relative efficacy of crowdsourcing applied to medical image analysis will be dependent on the complexity of the task, the papers examined here show crowdsourcing to be an effective methodology across a wide variety of applications, including objective assessment of surgical skill \citep{malpani2015study}, emphysema assessment \citep{orting2017crowdsourced}, polyp marking in virtual colonoscopy \citep{park2018crowd}, identification of chromosomes \citep{sharma2017crowdsourcing} and biomarker discovery in immunohistochemistry data \citep{smittenaar2018harnessing}. Notably, only one project stated that crowdsourcing could not always be applied effectively to the studied task (``it is very difficult and maybe even impossible to entirely outsource the task of labelling mitotic figures in histology images to crowds'' \citep{albarqouni2016aggnet}).

Rather than comparing the absolute performance of the crowd to experts or to algorithms, it might be worth considering their relative benefits. For example, crowds were particularly useful for rare classes~\citep{sullivan2018deep}, which are often difficult cases for algorithms. Another situation where crowds can be useful is identifying data that is missing from the gold standard provided by experts, see for example~\citep{luengo2012crowdsourcing}. Benefits of combining crowds with algorithms were also demonstrated by \citep{albarqouni2016aggnet,keshavan2018combining,sharma2017crowdsourcing}.

\subsection{Recommendations to ensure data quality}
The papers examined in this review included suggestions to improve the quality of data produced through crowdsourcing. These suggestions focused on refining the task design, crowdsourcing platform and post-processing of annotations. We summarize these recommendations here. 

\subsubsection{Task design} 
As discussed, crowdsourcing has been applied effectively to many medical imaging applications. However, careful study design remains necessary to ensure generation of data of sufficient quality. 

The selection and design of an appropriate crowdsourcing task is central to project success. Effort should be made to make the task simple and unambiguous \citep{rajchl2016learning,gurari2016investigating}, and to present study data appropriately \citep{mckenna2012strategies}. For unavoidably challenging tasks, crowdsourcing may still provide useful data, for instance, through enabling a rapid first-pass evaluation of large scale data sets \citep{della2014preliminary, park2017crowdsourcing}. Particularly challenging tasks may be made tractable through gamification \citep{albarqouni2016playsourcing} or careful reframing of the task, e.g. crowdsourcing of emphysema assessment was made possible through reframing the task as a question of image similarity \citep{orting2017crowdsourced}. Alternatively, it may be possible to achieve the desired data quality simply through asking a larger cohort of crowd workers to perform each task per data point.
An interesting example of task design is given in \citep{gurari2016investigating} where quality and speed of crowdsourced segmentations in natural images are increased by flipping images, suggesting that familiarity with an image can be detrimental.

\subsubsection{Crowdsourcing platform}
The choice of crowdsourcing platform  can influence study cost and completion time, as well as the size and demographics of the crowd. Furthermore, different platforms offer distinct features which may influence the quality of data produced. For example, \citeauthor{heller2017web} noted that user interface features, such as zoom and intuitive controls, can increase data quality. Contingent on the complexity of the task and interface design, training materials should be provided, as this can improve results \citep{mckenna2012strategies}. However, this is not always necessary - in some cases minimal \citep{brady2014rapid} or no training \citep{ganz2017crowdsourcing} was required.

\subsubsection{Post-processing}
Post-processing of annotations is recommended to improve annotation quality by removing annotations from poorly performing workers. Alternatively, if multiple workers annotate the same data it is possible to improve annotation quality by aggregating annotations.

The surveyed papers propose a variety of criteria for filtering individual annotations. For example, time spend on task \citep{oneil2017crowdsourcing}, expected shape of segmentation \citep{cheplygina2016early,chavez2013crowdsourcing}, correlation with other workers' results \citep{sharma2017crowdsourcing,chavez2013crowdsourcing} and correlation to experts annotations or ground truth \citep{sameki2016icord,keshavan2018combining, irshad2017crowdsourcing,irshad2015crowdsourcing,foncubierta2012ground}. However, due to the lack of comparisons between different filtering approaches, the only clear recommendation from these works is to use some form of filtering.

\citeauthor{nguyen2012distributed} found that filtering unreliable workers did not have a significant influence when annotations from multiple workers are aggregated. However, aggregating without taking individual performance into account might not be the best approach. \citeauthor{malpani2015study} compared different aggregation methods, and found that weighted voting, with weights based on self-reported confidence scores, improved results compared to simple majority voting. Similarly, \citeauthor{irshad2015crowdsourcing} found that aggregating segmentations from 3-5 workers, using weights based on consensus and worker trust scores, improved performance over using single worker annotations.
Further, \cite{cheplygina2018crowd} found that disagreement between workers was predictive of melanoma diagnosis in skin lesions, suggesting that simple aggregation, such as majority voting or mean statistics, might not be the best approach.

\section{Discussion}
\label{sec:discussion}

\subsection{Trends}

As discussed in Section~\ref{sec:applications}, crowdsourcing is applied to a variety of medical images, however, it is most commonly applied to histology or microscopy images. The trend for crowdsourcing of this image type may be due to the ease of which these (typically 2D) images can be incorporated into a crowdsourcing or citizen science project. Alternatively, the microscopy images examined in these papers may have not been derived from a patient, and would therefore not require the consent of an individual to use for crowdsourcing purposes. 

The most common crowd task is rating entire images. This is somewhat surprising, given that we would expect such tasks to rely more on prior knowledge than other crowdsourced tasks, such as drawing outlines of objects. Again, this trend might be facilitated due to the ease with which rating images can be incorporated in existing platforms.

Most crowdsourcing studies are set up on commercial platforms, followed by custom platforms. Each image is annotated by multiple crowd workers, who typically receive less than \$0.10 per annotation. On the one hand, this low reimbursement might be a product of researchers trying to optimize the total number of annotations given a particular budget. On the other hand, it could be a lack of awareness of what appropriate compensation should be \citep{hara2017data}.

A surprising finding is that, often, important details about the crowd and their compensation are missing. Besides missing details in terms of crowd compensation, we find missing details regarding the number of requested annotations per unit. While for some of the surveyed papers, we could infer an approximation of the number of annotations gathered per unit by checking the scale of the experiment and the total amount of annotations gathered, for at least a third of the surveyed papers (\pctAnnotatorsUnknown\%) this was not possible due to a lack of detail when describing the crowdsourcing experimental methodology.

Crowdsourced annotations are generally processed prior to evaluating how well the annotations solved the intended purpose. Simply excluding workers based on platform scores or a single test task is not as popular as continuously monitoring worker performance.
\pctAggregate\% of the surveyed papers aggregate annotations from multiple crowd workers. This is most commonly done by simple majority voting, but some papers use estimates of task difficulty and/or worker performance to obtain a weighted aggregation.

The most common approach to evaluating the quality of preprocessed annotations is by comparing to an expert defined gold standard. A smaller set of papers use the annotations to train a machine learning method and evaluate the performance of the trained method. 

The studies we reviewed almost unanimously conclude that crowdsourcing is a a viable solution for medical image annotation, which may seem unexpected given the complexity of medical imaging as a field in general. There might be several possible reasons for the lack of negative results. One is researchers selecting tasks which they already expect to be suitable for crowdsourcing. Another reason is publication bias, with papers demonstrating negative results having less chance of being published, which is also an issue in computer vision~\citep{borji2018negative}.

\subsection{Limitations}


There are a number of limitations in the way that the current studies are being conducted. There is generally a lack of clarity in the reporting of experimental design and evaluation protocols. Additionally, ethical questions regarding worker compensation, image content and patient privacy are rarely discussed, but seem crucial to address. 
In several papers the study design appears to be ad-hoc. Characteristics such as the platform, number of annotators, how the task is explained and so forth, are not always motivated, or even described. This creates difficulties in understanding what leads to a successful crowdsourcing study and increases the barrier for researchers who have not used crowdsourcing before. The studies which do examine such factors are often conducted on a single application, making it difficult to generalize lessons learned to other applications. Detailed documentation of experiments is a crucial factor for ensuring reproducible science and essential for replication studies.

Another problem is the evaluation of results. The quality of crowdsourced annotations is generally estimated by comparing directly to expert annotations. However, variation in both expert defined gold standard and crowd annotations are not systematically accounted for, making it difficult to assess if crowd annotations are actually good enough. When using annotations to train machine learning methods, noisy crowd annotations might not be a problem if handled by the method. However, variation in annotations should still be investigated in this case. A related problem is using expert annotations to filter crowd annotations, which would not be possible for real unlabeled data, thus leading to overly optimistic results. 

Overall, surveyed papers reported successful results. However, from our personal experience and discussions with other researchers, it is non-trivial to setup a crowdsourcing project for medical images. Due to the lack of negative results, the current literature does not inform researchers inexperienced with crowdsourcing about the main considerations of such a project. Furthermore, very few articles report on pilot experiments which aim to calibrate and identify the optimal crowdsourcing parameter settings such as the number of annotators per image.

There are important ethical issues which are largely not mentioned in the papers we surveyed. First of all, details about compensation are often missing, whereas this can have an important effect on the crowd~\citep{hara2017data}. Furthermore, what is reasonable compensation in one country, may be too low for another country due to different cost of living. How to set the compensation fairly is an open issue that researchers should consider in their work.  

Another ethical concern is whether it is possible and/or appropriate to share images with the crowd. Some images (for example of surgery) may be traumatic to view or unsuitable for children, which is more unique to the medical domain than other areas where crowdsourcing is applied e.g. astronomy or ecology projects. Another issue is sharing images from the perspective of patient consent, which is an issue that must be considered case by case.    

\subsection{Opportunities}


Several papers discuss directions they want to take in further research. One of the popular directions is increasing the role of machine learning. Several papers not using machine learning plan to do so in future \citep{brady2017improving,cheplygina2018crowd,sullivan2018deep}. Papers that already use machine learning discuss improvements to their algorithms or crowd-algorithm combinations \citep{sharma2017crowdsourcing,sameki2016icord}.


Related to the above, tailoring the tasks to individual workers is another possibility. The rating score given to workers by platforms only reflects an overall completion rate, and might be artificially high because employers tend to rate the majority of the tasks positive and apply a filtering afterwards. Considering worker scores on different task types could help to make a better selection of workers beforehand.

Another strategy discussed as future work is the use of gamification. Several papers have explored this idea~\citep{luengo2012crowdsourcing,mavandadi2012distributed,albarqouni2016playsourcing,sullivan2018deep} citing increased motivation of annotators. While the earlier papers~\citep{luengo2012crowdsourcing,mavandadi2012distributed,albarqouni2016playsourcing} have task-specific games, ~\cite{sullivan2018deep} takes a more task-independent approach of a mini-game within an existing, larger game. This could be an opportunity for many other researchers, without the need to design a game from scratch. Finally, annotating images at a  festival as in ~\citep{timmermans2016crowdsourcing} could be an interesting direction. 

Beyond the opportunities that the papers discuss as future research, we see a number of other future directions for the community as a whole. Perhaps the most important future direction is openly sharing our experiences with crowdsourcing, including failures. Due to publication bias, current papers may not reflect the performance and difficulties encountered in a typical crowdsourcing project. 

More generally, there is an opportunity to create a set of guidelines for crowdsourcing medical imaging studies. Rather than relying on ad-hoc choices, researchers could then make informed decisions about the platform, reward of the annotators and other variables. For example, the European Citizen Science Initiative has a selection of guides for performing citizen science studies\urlfootnote{https://ecsa.citizen-science.net/blog/collection-citizen-science-guidelines-and-publications}. A further opportunity is to interact more with other fields where crowdsourcing has been in use longer, and to see which of their best practices are also applicable to medical imaging. 

Interacting with workers could both improve projects and help establish guidelines. Workers have created communities (e.g. Reddit, Facebook) and discussion boards (\url{https://www.mturkforum.com}, \url{https://www.turkernation.com}) for some platforms. \citeauthor{chandler2014nonnaivete} found that 28\% $\pm$ 5\% of the workers on Mechanical Turk read discussion boards and blogs related to Mechanical Turk. The topics of conversations, in order of frequency, are: pay, gratification, completion time, difficulty, how to successfully complete, purpose and the requesters' reputation.
These forums are a valuable source for researchers for gathering information, measuring opinions and getting feedback on improving their project. This is particularly important because high throughput workers are more likely to discuss taskss~\citep{chandler2014nonnaivete}. This subgroup (less than 10 \% of the workers do more than 75\% of the work \citep{hara2017data}) is likely to have experience with similar tasks \citep{chandler2014nonnaivete}, and interaction with these workers may result in various improvements such as improvements of the user interface as in \citep{bruggemann2018exploring}.

Next to image analysis, crowdsourcing could also be a way to collect, rather than curate, data to improve medical knowledge. This could vary from donating your own medical images (such as \url{http://www.medicaldatadonors.org}) to contributing experiences about rare diseases. Since such initiatives do not focus on image analysis we did not include them in this survey, however ~\citep{ranard2014crowdsourcing,wazny2017crowdsourcing} may be good starting points for readers interested in these topics.

\section*{Acknowledgments}

The authors would like to thank eScience-Lorentz grant 2018 and Ms Gerda Filippo (Lorentz center) for their support in organizing the workshop where this paper was conceived. We thank the researchers who responded to our preprint for their valuable comments and suggestions.    

Silas {\O}rting was supported by the Danish Council for Independent Research (DFF) and  the Netherlands Organization for Scientific Research (NWO).

Matthias Hirth was supported by Deutsche Forschungsgemeinschaft (DFG) under Grants HO4770/2-2, TR 257/38-2. The authors alone are responsible for the content.

Oana Inel was supported by the IBM PhD Fellowship program.

\section*{Appendix A - Data}

Our summaries of the surveyed papers - a longer summary and the short summary presented in Table \ref{tab:literature} - are available for download via Figshare, \url{http://tiny.cc/9gxybz}.

\bibliography{refs_veronika,refs_public}
\end{document}